# Towards Network Traffic Monitoring Using Deep Transfer Learning

Harsh Dhillon and Anwar Haque
Department of Computer Science
Western University
London, Ontario, Canada
{hdhill48,ahaque32}@uwo.ca

*Abstract*—Network traffic is growing at an outpaced speed globally. The modern network infrastructure makes classic network intrusion detection methods inefficient to classify an inflow of vast network traffic. This paper aims to present a modern approach towards building a network intrusion detection system (NIDS) by using various deep learning methods. To further improve our proposed scheme and make it effective in real-world settings, we use deep transfer learning techniques where we transfer the knowledge learned by our model in a source domain with plentiful computational and data resources to a target domain with sparse availability of both the resources. Our proposed method achieved 98.30% classification accuracy score in the source domain and an improved 98.43% classification accuracy score in the target domain with a boost in the classification speed using UNSW-15 dataset. This study demonstrates that deep transfer learning techniques make it possible to construct large deep learning models to perform network classification, which can be deployed in the real world target domains where they can maintain their classification performance and improve their classification speed despite the limited accessibility of resources.

## I. INTRODUCTION

Internet Service Providers (ISP) globally have witnessed a fast expansion in their network traffic over the last few decades. Our societies are now closely intertwined with various networking services to perform its many day-to-day functions. Promising technologies such as 5G are currently making headway to increase the speed of connections between our machines swiftly. This trend will continue to evolve our network communication systems to become much rapid with each passing year. With so much reliance on communication networks, the infrastructure becomes a key target for cybercrimes, which can now have a more significant impact. Network security systems also need rapid improvements as we cannot rely on classic intrusion detection and scanning techniques, which have become quite obsolete to give accurate classifications in a dynamic and massive volume network traffic scenario.

Deep learning models excel at learning from a large dataset of labeled examples. Training a sizeable deep learning model also requires significant computational resources. Over the last decade, a lot of research and effort has been given to the space of supervised learning. In the real-world domain, supervised models usually generalize well if the environment is similar to the one where the end-to-end model was trained in as it expects similar data and computational resources to maintain its performance. The real world is non-deterministic and can present an infinite number of possibilities and patterns unseen by the deep learning algorithm. To navigate through such scenarios and perform with ideal accuracy and speed, learning models will need a prior understanding of the tasks and related domain at hand.

In this paper, we propose and demonstrate a novel deep transfer learning-based Intrusion Detection System architecture, which uses the prior knowledge of the models trained on larger datasets to perform at high speed while maintaining an optimal accuracy, despite the low availability of both data and computational resources. We will use the current USNW-15 dataset to demonstrate the proposed architecture. This research experimentally showcases that integrating transfer learning techniques in the core design of Intrusion Detection Systems can improve their overall efficacy in real-world settings. The novel methods outlined in this paper enable the IDS to utilize large and powerful deep learning models without the need for high computing power and extensive data. Using the defined methods, we can also effectively boost the classification speed of an IDS model, which enables it to maintain its accuracy while performing at real-time processing speeds.

This paper is organized as follows. Section II describes the historical background of modern intrusion detection systems and related work in the field of machine learning and deep learning. Section III describes the foundational concepts of transfer based learning. Section IV presents the proposed methodology to architect the deep transfer learning-based IDS. Section V presents the benchmark performance results of the applied deep learning algorithms in the architecture. Section VI concludes our paper and outlines the ideas for future research undertakings.

## II. RELATED WORK

The earliest sketch of a real-time intrusion detection system was proposed by Dorothy E. Denning in 1986 [1]. Her work was inspired by the prior study of Jim Anderson in 1980, which formulated a way to audit a computer's data to identify abnormal usage patterns at the end of each day using a statistical analysis approach [2]. This research further augmented into IDES, abbreviated for Intrusion Detection Expert System developed by Teresa F. Lunt at SRI International in 1988 [3]. IDES had two main components. The first component adaptively learns the user's normal behavior pattern and detects patterns that deviate from them. The second component uses a rule-based approach to encode the encountered system vulnerabilities and store them in a knowledge base. Lunt proposed integrating an artificial neural network in the expert system as a third component, which was not fully implemented in the follow-up derivations of IDES. By the 1990s, intrusion detection systems were started to get implemented by various research labs and business computing firms, including AT&T Bell Labs, who built their own versions of detection systems, using IDES as a base on various other hardware and different programming languages [4]. The introduction of well-labeled KDD-99 intrusion detection dataset enabled researchers to work in the field of computer security to apply data mining and machine learning algorithms to build much efficient and generalized IDS [5]. In 2001, Tamas Abraham used data mining techniques to formulate the IDDM, abbreviated for Intrusion Detection Using Data Mining architecture [6]. Traditionally, data mining systems operated on large off-line data sets. IDDM architecture was designed to use data mining in real-time environments to identify anomalies and misuse. Li Jun et al. proposed a hierarchical network intrusion detection system, which used a Perceptron-Backpropagation hybrid model to classify anomalous and normal network traffic to recognize UDP flood attacks [7]. In 2002, Eskin et al. proposed an unsupervised intrusion detection framework using SVM, K-Nearest Neighbor, and clustering algorithms [8]. Weiming Hu et al. used an Adaboost-based algorithm with an adaptive weight strategy to build a detection model reporting low computational complexity and error rates [9]. J. Zhang et al. used random forest algorithm-based data mining techniques to build a hybrid IDS, which is capable of functioning as both a misuse and anomaly detection system [10]. Chandrasekhar et al. applied fuzzy neural networks to build their variation of IDS [11], which claimed better experimental results than the Backpropagation Neural Networks and other well-known machine learning methods.

The 2012 ImageNet victory led by Hinton et al. demonstrated that deep neural networks were able to outperform complex machine learning models in image recognition tasks [12] by beating the state-of-the-art algorithms by a whopping 10.8 percentage point margin rate and creating a renewed interest in the field of deep learning. In the proceeding years, academics working in computer security also started integrating deep neural networks in their research. In 2014, Wang et al. applied deep belief networks, a class of DNN, which reported the lowest published false-positive results with the KDD-99 dataset [13]. N. Moustafa et al. [14] reinvigorated the field with their UNSW-NB15 network data set, which is much superior for evaluating NIDS performance, as it reflects the modern traffic scenarios more fittingly than decade-old intrusion datasets such as KDD-99 and NSLKDD. In 2018, N. Moustafa et al. used the UNSW-NB15 dataset to create NIDS for IoT traffic data for classifying normal and suspicious instances by applying AdaBoost ensemble techniques [15]. A. Ahim et al. [16] combined different classifier approaches based on decision trees and various rules-based concepts to build a novel IDS using the CICIDS2017 dataset. In 2019, Y. Xiao et al. [17] implemented a CNN based IDS using Batch Normalization with KDD99 Dataset, the demonstrated CNN-IDS displayed fast classification times than non-CNN algorithms, making a deep neural network approach ideal to construct a real-time IDS. Vinayakumar et al. [18] created a hybrid IDS to monitor network and host level activities. Upon conducting an exhaustive comparative study with various machine learning and deep learning classifiers, DNN was demonstrated to outperform other applied classifiers. B. Riyaz et al. [19] designed an IDS using deep learning approach in wireless networks using the combination of novel coefficient-based feature selection algorithm (CRF-LCFS) and a CNN using KDD-99 dataset. Their proposed method demonstrated a 98.9% detection accuracy. In our research, we will be using a hybrid CNN-LSTM model to train our IDS to filter network traffic in a source domain at a high classification accuracy. Using transfer learning techniques, we will transfer the domain knowledge learned by our model in a simulated real-world environment. Our experiments demonstrate that using the novel method outlined in this paper, the applied unified model improves its classification accuracy in the real world target domain as well as efficiently increases its performance speed.

III. TRANSFER LEARNING

Transfer learning is a concept where a learning algorithm reuses the knowledge from the past related tasks to ease the process of learning to perform a new task [20]. The ability to transfer the knowledge gained from previous tasks has a wide range of real-world applications, including building real-time intrusion detection systems that can perform optimally even with scarcity of data and computing resources. Using deep transfer learning alleviates the massive data dependency of deep learning algorithms, which they require to learn the underlying patterns in the data. In general terms, using transfer learning, we aim to transfer the knowledge from a source domain to a target domain by relaxing the assumption that the training data and the test data must be independent and identically distributed, which is rare for the real-world data. Fig. 1 shows the process of transferring a model's network architecture and learned weights from a source domain with large dataset and higher computational resources to a target domain with a smaller dataset and limited computational resources.

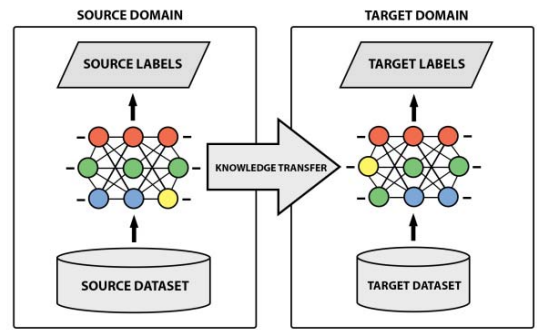

Fig. 1. Transfer Learning Concept

A *domain* can be represented as, $D = \{X, P(X)\}$, which consists of two parts: the feature space $X$ and a margin distribution $P(X)$, Where $X = \{x_1, x_2, \ldots, x_n\}, x_i \in X$. Whereas A *task* can be represented as, $T = \{Y, P(Y|X)\} = \{Y, \eta\}$, $Y = \{y_1, y_2, \ldots, y_n\}, y_i \in Y$, where $Y$ is a label space, and $\eta$ represents the predictive function which can be learned from the training data including pairs $\{x_i, y_i\}$, where $x_i \in X, y_i \in Y$; for each feature vector in the domain, $\eta$ predicts its corresponding label as $\eta(x_i) = y_i$[21].

In this paper, we consider our source domain as $D_S$, and target domain as $D_T$. The source domain data is denoted as $D_S = \{(x_{S_1}, y_{S_1}), \ldots, (x_{S_n}, y_{S_n})\}$, where $x_{S_i} \in X_S$ is the data instance and $y_{S_i} \in Y_S$ is the corresponding class label. In our IDS, $D_S$ is the set of term vectors together with their associated attack and malicious labels. Similarly, we denote the target domain data as $D_T = \{(x_{T_1}, y_{T_1}), \ldots, (x_{T_n}, y_{T_n})\}$, where the input $x_{T_i}$ is in $X_T$ and $y_{T_i} \in Y_T$ is the corresponding output [21]. We can now give the transfer learning definitions as follows,

Given a source domain $D_S$, learning task $T_S$, a target domain $D_T$ and learning task $T_T$, transfer learning aims to help improve the learning of the target predictive function $\eta_t$ by using the knowledge in source domain $D_S$ and learning task $T_S$, where $D_T \neq D_S$, or $T_S \neq T_T$. The size of $D_S$ is much bigger than $D_T$ in various applied situations.

Additionally, when there exists some relationship, explicit or implicit, between the feature spaces of the two domains, we say that the source and target domains are related. In this paper, the two domains are related as they share a similar feature space from intrusion datasets. A transfer learning task defined by $(D_S, T_S, D_T, T_T, \eta_t)$ becomes a deep transfer learning task if $\eta_t$ is a non-linear function represented by a deep neural network.

Chuanqi Tan et al. [21] classified deep transfer learning approach into four main categories, namely instance-based, mapping-based, network-based, and adversarial-based transfer learning. In this paper, we utilize the network-based transfer learning approach. Network transfer learning refers to the transfer of partial network trained in the source domain, which includes its network structure and learned weights to the target domain, where it becomes the part of its existing architecture. The network-based transfer learning architecture works with the notion that the neural networks should become as iterative as human brains. Human brains use prior knowledge even when they are performing new tasks and often perform well with the new tasks by using the previously learned concepts.

## IV. PROPOSED METHODOLOGY

### A. Database

For architecting our transfer learning-based IDS model, we will use the USNW-15 dataset. This dataset is relatively modern when compared to other widely used datasets like KDD99 and NSL-KDD in network security research. USNW-15 was created at Cyber Range Lab of the Australian Centre for Cyber Security (ACCS). The dataset contains nine types of malicious attacks, namely Analysis, Backdoors, DoS, Exploits, Fuzzers, Generic, Reconnaissance, Shellcode, and Worms. In total, the UNSW-15 dataset contains 100 GB worth of raw network packet observations. We will use a partition set of this data which includes 257,673 records and will further divide the selected partition into a training set with 154,603 records. We will also use a validation set and a testing set, both with 51,535 records, respectively, to aptly evaluate the performance of the applied deep learning models in the separate domains. The raw packet data in the UNSW-15 dataset was recorded over the modern network infrastructure, which helps in building an IDS that can generalize well in the real-world environment as well. We will only use the training set and validation set in our source domain for training and validating our learning models and to benchmark their performance. In our target domain, we will use the testing set, which is the unseen partition of data, which simulates the real-world unobserved data for our models to test their performance.

### B. Feature Engineering

The features we select to a model the IDS architecture form the core of the classifier, aiming to infer and differentiate between normal and malicious packets. USNW-15 dataset has in total of 49 features. To optimize our system further, we made feature selection to choose features that are substantial to the classification task.

TABLE I
DATASET KEY FEATURE DESCRIPTION

| Feature | Description |
|---|---|
| sload | Source bits per second. |
| dload | Destination bits per second. |
| stcpb | Source TCP base sequence number. |
| dtcpb | Destination TCP base sequence number. |
| sbytes | Source to destination transaction bytes. |
| dbytes | Destination to source transaction bytes. |
| sttl | Source to destination time to live value. |
| dttl | Destination to source time to live value. |
| swin | Source TCP window advertisement value. |
| dwin | Destination TCP window advertisement value. |
| sjit | Source jitter (millisecond). |
| djit | Destination jitter (millisecond). |
| spkts | Source to destination packet count. |

We use feature importance to judge each feature based on a scoring value, which represents how important and relevant the feature is to the output variable and vice versa. We used a tree-based extra-trees classifier, which is an ensemble model of randomized decision trees to identify and select the best features in the dataset. Table I describes the most significant features selected during the feature selection step. Based on the results from the tree-based feature importance technique, we dropped 15 features from our dataset, which had the least scoring performance. We observed that filtering the dataset using such methodology improved the overall accuracy and speed performance of the applied deep learning techniques.

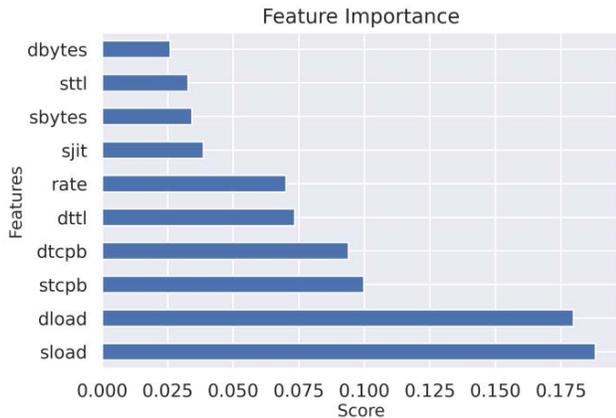

Fig. 2. Feature Importance Bar Chart

Fig. 2 shows the top ten identified features in the dataset. Feature selection enables the model to allocate its computational resources appropriately, which also leads to an increase in the training time because we are reducing the data to process and construct the model. The presence of irrelevant and redundant data makes the ultimate goal of knowledge discovery much harder as well.

### C. Data Normalization

Data normalization or feature scaling is a data preprocessing technique where we convert all input values used in the learning model to a common scale. Without scaling the data present in the dataset, the features with large value will have a greater impact on the output of the learning model. Such scale difference leads to important features with smaller range become less effective to the overall inferences drawn by the classifier. To make all the features equal, it is important to normalize or scale our data, which also helps the algorithm reach convergence faster. Fig. 3 visualizes how normalization changed the natural range of raw feature named dbytes to a standard range of [0,1] as an example.

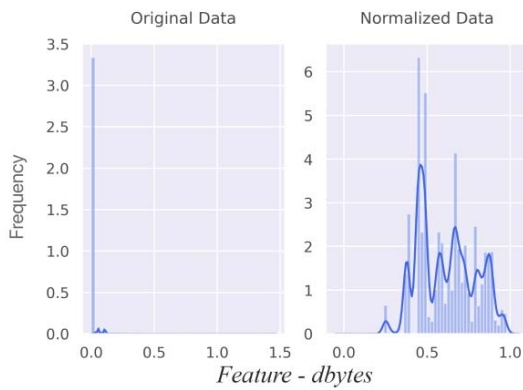

Fig. 3. Data Normalization Visualization

Normalization independently rescales the data feature-wise from its natural range into a standard range where for every feature the minimum value gets transformed into the value of zero, and the maximum value gets transformed into the value of one, hence giving all the features in data an equal footing for drawing the statistical inference. We used min-max normalization whose formula can be expressed as,

$$\acute{x} = \frac{(x - x_{\min})}{(x_{\max} - x_{\min})}$$

where $x$ represents the scaling data point, $x_{\min}$ is the minimum and $x_{\max}$ is the maximum absolute value of $x$. The min-max normalization retains the shape of the feature intact during scaling which helps the model avoid overfitting as compared to other normalization techniques we tested during our experiments. This particular data pre-processing step is vital as various algorithms such as logistic regression and neural networks etc. assume that the input data for drawing the inference from will be duly scaled and normalized.

### D. Learning Algorithms

Deep learning algorithms are capable of achieving higher accuracy in terms of classification when compared to other shallow networks and machine learning models. After training the chosen deep learning algorithm in the source domain, we use transfer learning methodology to transfer the model's architecture and learned weights to a target domain and test the performance of the model in the new domain with the unseen dataset partition. To choose the deep learning algorithm for our source domain architecture, we did a benchmark study on three deep learning algorithms, namely Deep Neural Network (DNN), Convolutional Neural Network (CNN), and Long Short-Term Memory architecture fitted with Convolutional Neural Network in its hidden layers (CNN-LSTM). The three types of learning algorithms are briefly summarized as follows.

A Deep Neural Network consists of multiple fully-connected layers, which passes information from numerous layers in a feed-forward manner such as the input layer, several hidden layers, and an output layer to learn its weights iteratively using backpropagation algorithm, which computes the partial derivatives of the cost function with respect to each neuron unit with respect to its weights and biases to reach the local minima of the function. LeCun et al. [22] demonstrated one of the earliest practical implementations of backpropagation algorithm to build a handwriting recognition OCR.

Convolutional Neural Network is a class of deep neural networks which are applied to image recognition tasks as they can learn highly representative and hierarchical features from their input. Each layer inside a CNN is composed of several neuron units. The neurons are organized in a style that the output of neurons at layer $l$ becomes the input of neurons at layer $l + 1$, such that

$$a^{(l+1)} = f(W^{(l)}a^{(l)} + b^{(l)})$$

where $W^{(l)}$ is the weight matrix of layer $l$, $b^{(l)}$ is the bias term, and $f$ represents the activation function. The activation for layer $l$ is denoted by $a^{(l)}$. A CNN consists of a convolutional layer which extracts the features from an input vector using filters, a ReLU unit for introducing non-linearity in the network, a pooling layer which reduces the dimensionality of the data while keeping the useful information and finally a fully connected layer which computes the potential output using a softmax function.

Long Short-Term Memory networks are a variant of recurrent neural networks, which in addition to feedforward connections, also have looping feedback connections that allow the model to store persistent information over a period of time. LSTM first proposed by S. Hochreiter et al. [23], are capable of learning long-term and short-term dependencies without losing or over accumulating information. LSTM is capable of adding or removing information in their cell states by using regulation structures called gates, which control the flow of data for each cell unit present in its architecture. Such design gives LSTM an advantage over conventional feed-forward neural networks because of their ability to selectively retain or drop information.

### E. Proposed IDS Model

In this paper, our chosen deep learning architecture for the IDS consists of a CNN with LSTM present in its hidden layers and fully connected layer units to predict the classification labels. As shown in Fig. 3, the proposed unified IDS model can use the advantages of the three distinguished deep learning models and combines their latent feature extraction, memory retention, and classification abilities to give a higher accuracy score as compared to the models applied separately.

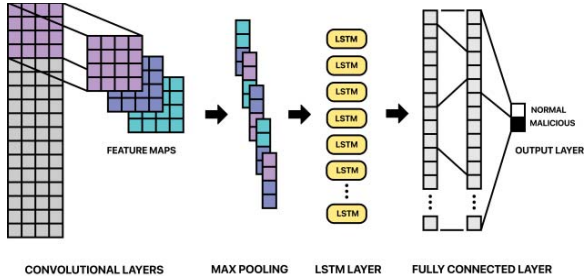

Fig. 3. IDS Learning Model

A CNN is capable of learning and recognizing patterns over an input space, whereas LSTM units can learn and recognize patterns across time. A DNN or a fully connected layer, on the other hand, is capable of learning mappings from an input vector to give precise class wise outputs. Both CNN and DNN belong to the class of feedforward networks where data can only flow in the forward direction. CNN can use a 2D input and transform it into internal vector representations to further extract its features. In contrast, when we apply LSTM with CNN, LSTM provides the capability of using the feature vector output of the CNN and further build internal states whose weights can repeatedly be updated because data in LSTM flows in a recurrent manner. During this entire process, the CNN extracts the inherent features from the input. In contrast, LSTM interprets those features across various time steps, which makes the architecture more efficient to learn deeper representations and relationships in the data, in contrast with any network architecture applied separately.

Combining DNN, CNN and LSTM have been explored in the past in [24], where the models are being trained separately, and then their outputs are later combined. In our approach, we are training the unified model jointly with each model providing their processed feature outputs as an input to the subsequent models in the scheme.

Table II shows the summary of our candidate CNN-LSTM model, where we are first using CNN layers to extract the contextual features in the training set. The utility of CNN's to downsample the input while conserving the important features during the extraction process reduces the overall dimension of the feature parameters. The output of CNN is then fed into the LSTM layers to model the signal in time and train the weights using backpropagation in time (BPTT) algorithm. Finally, after the signal is modeled in the LSTM layers, the output is passed into fully connected layers, which are used to learn higher-order feature representations that are suitable for separating the output into different class labels.

TABLE II
CNN-LSTM IDS MODEL ARCHITECTURE

| Layer Type | Output Shape | Total Units |
| --- | --- | --- |
| conv1d_1 (Conv1D) | (None, 32, 64) | 256 |
| conv1d_2 (Conv1D) | (None, 32, 64) | 12352 |
| max_pooling1d_1 (Pooling) | (None, 16, 64) | 0 |
| conv1d_3 (Conv1D) | (None, 16, 128) | 24704 |
| conv1d_4 (Conv1D) | (None, 16, 128) | 49280 |
| max_pooling1d_2 (Pooling) | (None, 8, 128) | 0 |
| conv1d_5 (Conv1D) | (None, 8, 256) | 98560 |
| conv1d_6 (Conv1D) | (None, 8, 256) | 196864 |
| max_pooling1d_3 (Pooling) | (None, 4, 256) | 0 |
| lstm_1 (LSTM) | (None, 100) | 142800 |
| dense_1 (Dense) | (None, 256) | 25856 |
| dropout_1 (Dropout) | (None, 256) | 0 |
| dense_2 (Dense) | (None, 128) | 32896 |
| dropout_2 (Dropout) | (None, 128) | 0 |
| dense_3 (Dense) | (None, 1) | 129 |

### F. Experiments Setup

For our experimentation, we provisioned a VM cluster in the Google Cloud Platform (GCP). We used the VM cluster instance type n1-standard-16, which was configured with 16 vCPUs and a 30GB RAM allocation. For our deep learning libraries, we used Keras Framework with TensorFlow1.15 in the back-end. For data-preprocessing and manipulation, we used Pandas and Scikit-learn libraries. The experiments were ran using Jupyter notebook IDE and Python 3.7. To further simulate a resource sparse target domain, we used another provisioned VM in GCP. We used the VM cluster instance type n1-standard-1, which is configured with one vCPU and 3GB RAM allocation.

*G. Performance Evaluation*

To evaluate the performance of our applied model, we will use measures namely Classification Accuracy, Confusion Matrix and ROC curve briefly described as follows,

a. Classification Accuracy is a metric used for classification models, where we compare the number of correct predictions drawn by the model with the total number of predictions made by the model. The classification accuracy can be expressed as,

$$Accuracy = \frac{Number\ of\ Correct\ Predictions}{Total\ Number\ of\ Predictions} * 100$$

b. Confusion Matrix is a visual representation for the performance of a classification model where the outcome of the model is expressed using four key categories. True Positive refers to the values which were predicted to be positive, and they are indeed positive and hence true. False-positive refers to the values which were predicted to be positive but are negative and thus false. True Negative refers to the values which were predicted to be negative and indeed are negative and hence true. False Negative refers to the values which were predicted to be negative but are, in fact, positive and thus false.

c. Receiver Operating Characteristic (ROC) curve, is a curve plot where we compare two parameters, True Positive Rate and False Positive Rate. True positive rate, also known as sensitivity of the model, determines the proportion of values that are positive and were correctly identified as positive by the model. TPR can be expressed as,

$$TPR = \frac{True\ Positive}{True\ Positive + False\ Negative}$$

The false-positive rate, also known as specificity of the model, determines the proportion of the values which are negative and were also identified as negative by the model. FPR can be expressed as,

$$FPR = \frac{False\ Positive}{False\ Positive + True\ Negative}$$

ROC curve plots the True positive rate of a model with respect to its False positive rate at various thresholds.

## V. RESULTS

*A. Source Domain Architecture*

For selecting the candidate source domain IDS model, we performed a benchmark study on three deep learning models, namely DNN, CNN, and CNN-LSTM. Our results show that CNN architecture demonstrated a 92.16% accuracy score on the source domain validation dataset, whereas DNN architecture demonstrated an accuracy score of 87.66%. The CNN architecture with LSTM layers present in its hidden layers demonstrated a 98.30% accuracy score outperforming other applied models. Fig. 4 shows the bar plot of model accuracies. Based on these results, we chose CNN-LSTM model as our candidate model for IDS architecture.

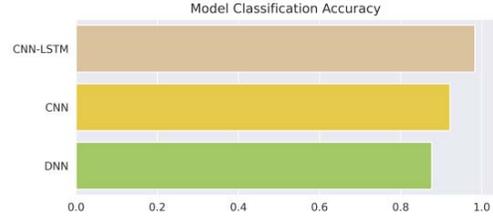

Fig. 4. Accuracy Score – Source Domain

The results show that CNN-LSTM was able to learn more representational features in the training data and was able to generalize well to the validation dataset. As discussed previously, combining the three model architectures enhanced the overall classification accuracy. DNN is suitable for the task for generating higher-order feature representations, which can be separated into distinctive classes, but they don't enforce any structure or local information in the data. CNN, on the other hand, is suitable for extracting important features by condensing the input data to find the inherent structures and representations in the data, which gives a performance edge over DNN for the packet classification task. Adding LSTM units in hidden layers of CNN further improves the architecture. With the utility of LSTM, the output being parsed from CNN can further be modeled temporally using recurrence before feeding the results into DNN, which is vital for classifying the features into their appropriate class labels.

Accuracy by itself may not be the best evaluation metric for performance. Hence, we used confusion matrix for CNN-LSTM architecture to study the classification results in-depth, as shown in Fig. 5.

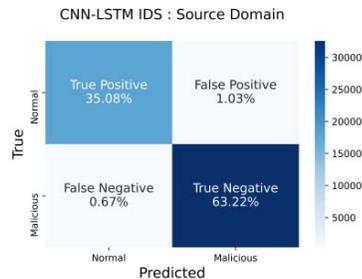

Fig. 5. Confusion Matrix – Source Domain

As per the confusion matrix, the CNN-LSTM model demonstrates a 1.03% false-positive rate and an 0.67% false-negative rate. False classifications have been a major area of concern for IDS, as systems can incorrectly classify malicious packets as normal ones leading to successful intrusion scenarios that can debilitate the infrastructure. The normal packets can also be classified as malicious packets, which leads to false alarms and the dropping of good packets, which might be useful, reducing the overall quality of service of the network.

The applied models were also evaluated using the ROC metrics, as shown in Fig. 6.

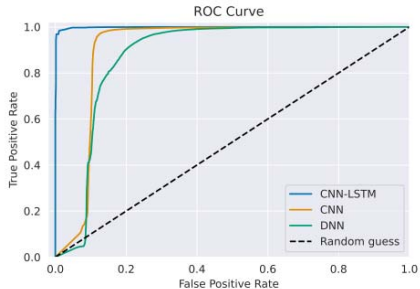

Fig. 6. ROC Curve – Source Domain

*B. Target Domain Architecture*

To apply the learned knowledge in the target domain, we will use the source domain's CNN-LSTM model architecture as well as its learned weights and transfer them to the model in our target domain. We will use the unseen testing data-set in this domain to simulate the IDS model being in a real environment where it encounters new data. This helps in evaluating how the model will essentially react when it is deployed in a real-world network infrastructure.

Our experiments show that using the learned weights from the source domain improved the overall performance and speed of models in the target domain. The DNN architecture demonstrated an improved accuracy score of 88%, whereas CNN architecture reported a 91.88% accuracy score. Our candidate model CNN-LSTM architecture demonstrated an accuracy of 98.43%, which is higher than other applied models in this domain. The Fig. 7 shows the bar plot of accuracy score in the target domain.

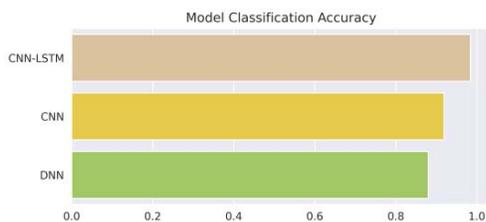

Fig. 7. Accuracy Score – Target Domain

The CNN-LSTM model also demonstrated better binary classification results with a general reduction in the false positive and false negative rates. Overall, the model has a 0.95% false-positive rate and a 0.62% false-negative rate, improving the packet classifications, as shown in Fig. 8.

To further study our results in the target domain, we used ROC curve metrics plotted in Fig. 9, which is noted to be comparable to the source domain ROC. It reflects the maintenance of the classification ability by the applied model with the unseen testing dataset in the target domain.

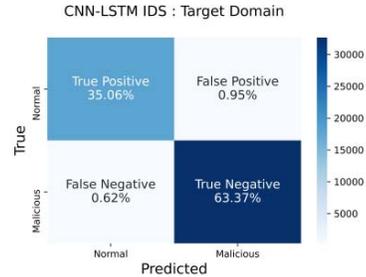

Fig. 8. Confusion Matrix – Target Domain

In essence, because we are using the transfer learning methodologies, the model is not being trained from scratch in the target domain. We are applying the knowledge learned prior to the source domain to perform the task of packet classification again in the target domain. Since the model has learned this task beforehand, it becomes easier for it to perform the same task again in the target domain. Various weights parameters between numerous units present in the models are not being modified; hence we don't need a large training set to retrain the model again. The low computational resources in the target domain are enough for the model to work with reasonable efficiency on the unseen data. We also observe that the knowledge transferred from the source domain improved the classification ability of IDS in the target domain. Overall, the testing speed of the applied deep learning models improved as well by a large margin in the target domain. DNN had the fastest testing speed in both domains but gave a low accuracy score because it is a simpler feedforward model with limited hierarchical feature extraction ability. CNN-LSTM had the slowest testing speed due to the inherent complexity of its architecture as well as the presence of more units and parameters as compared to both DNN and CNN models, but the model had a better accuracy score and classification results.

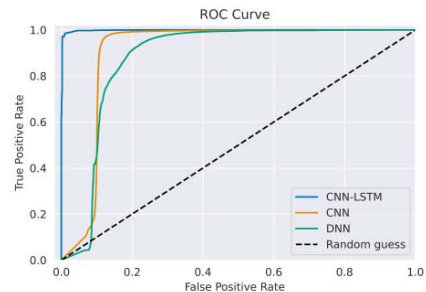

Fig. 9. ROC curve – Target Domain

TABLE III
MODEL ACCURACY AND TEST SPEED

| DL Model | Source Domain | | Target Domain | |
| --- | --- | --- | --- | --- |
| | Accuracy | Test Speed | Accuracy | Test Speed |
| DNN | 87.66% | 32.8s | 88.00% | 1.67s |
| CNN | 92.16% | 134.2s | 91.88% | 18.4s |
| CNN-LSTM | 98.30% | 189.5s | 98.43% | 22.3s |

Table III shows the summarized results of our benchmark study with three learning algorithms, namely DNN, CNN, and CNN-LSTM. From our results, we observed that CNN-LSTM outperformed other applied models by giving more accurate classification results with a 98.30% accuracy score in the source domain and a 98.43% accuracy score in the target domain. The maintenance of accuracy reflects the model's ability to utilize the weights learned from the source domain and apply them in the target domain to generalize well on an unseen testing dataset. The testing speed also improved despite the target domain's simulated data and computational resource scarcity.

## VI. Conclusion and future work

Our study demonstrates that deep transfer learning approach can be highly effective in developing an efficient, unified network intrusion detection system that maintains and improves its classification accuracy and speed in a simulated real-world setting via knowledge transfer. Using the proposed method, we can train a large and powerful deep learning IDS model in a source domain with a high allocation of data and computational resources. After validating our model's performance, we can then transfer its architecture and learned weights in a target domain with reduced computational resources, where we observe that the model maintains its efficiency as well as improves its testing speed. The target domain is aimed at simulating the real-world environment where we are using a partition of the dataset, which is entirely unseen by our models during their training and development.

This research showcases that high powered deep learning based IDS architectures can be deployed on real-world devices with lesser resources, which can maintain their efficiency and improve their speed using the transfer learning approach. Applying transfer learning in the overall design of an IDS not only enhances its performance in a real-world setting but also essentially increases its speed of classification, which is a tremendously required feature demanded by an IDS to protect and secure modern network infrastructures. Our research is one of the earliest practical implementations of integrating transfer learning techniques in the core architecture of an IDS.

As future work, we would like to integrate dimensionality reduction techniques in the existing IDS architecture and use different modern network intrusion datasets to further validate the proposed model and techniques.